\documentclass[10pt,twocolumn,letterpaper]{article}

\usepackage{cvpr}
\usepackage{times}
\usepackage{epsfig}
\usepackage{graphicx}
\usepackage{amsmath}
\usepackage{amssymb}

\usepackage{epsfig}
\usepackage{amsmath}
\usepackage{amssymb}
\usepackage{color}
\usepackage{comment}
\usepackage{bbm}
\usepackage[ruled,lined,boxed,linesnumbered]{algorithm2e}
\usepackage{algorithmic}
\usepackage{multirow}
\usepackage{rotating}
\makeatletter
\@namedef{ver@everyshi.sty}{}
\makeatother
\usepackage{tikz}
\usepackage{tikzscale}
\usepackage{tikz-qtree,tikz-qtree-compat}
\DeclareUnicodeCharacter{0301}{\'{e}}
\newcommand{\off}[1]{}

\newcommand{\bx}{\mathbf{x}}
\newcommand{\by}{\mathbf{y}}

\newcommand{\bw}{\mathbf{w}}

\newcommand{\bdiff}{\mathbf{diff}}

\newcommand{\real}{\mathbb{R}}

\newcommand{\TV}{{V^\beta}}

\usepackage[pagebackref=true,breaklinks=true,letterpaper=true,colorlinks,bookmarks=false]{hyperref}

\cvprfinalcopy % *** Uncomment this line for the final submission

\ifcvprfinal\fi
\begin{document}

%%%%%%%%% TITLE
\title{What do CNN neurons learn: Visualization \& Clustering}

% \author{Haoyue Dai, Zichao Yang, Xiao Li \\
%         Shanghai Jiao Tong University \\
%         {\tt\small \{markdai, sjtuyangzichao, shjdlx\}@sjtu.edu.cn}}

\author{Haoyue Dai\\
        Shanghai Jiao Tong University\\
        {\tt\small markdai@sjtu.edu.cn}}

\maketitle
%\thispagestyle{empty}

%%%%%%%%% ABSTRACT
\begin{abstract}
In recent years convolutional neural networks (CNN) have shown striking progress in various tasks. However, despite the high performance, the training and prediction process remains to be a black box, leaving it a mystery to extract what neurons learn in CNN. In this paper, we address the problem of interpreting a CNN from the aspects of the input image's focus and preference, and the neurons' domination, activation and contribution to a concrete final prediction. Specifically, we use two techniques – visualization and clustering – to tackle the problems above. Visualization means the method of gradient descent on image pixel, and in clustering section two algorithms are proposed to cluster respectively over image categories and network neurons. Experiments and quantitative analyses have demonstrated the effectiveness of the two methods in explaining the question: what do neurons learn.
\end{abstract}

\section{Introduction}

Nowadays CNN have obtained satisfying progress in research fileds and yielded promising application scenarios in industry. However, despite the excitement in perfomance, due to the end-to-end method in both training and prediction processes, CNN seems to be a black box: we know it indeed learns something to be effective, but we don't know why or what do the neurons learn~\cite{learnquestion}. This basic question forms the significant challenge in CNN interpretabily research: black boxes always tend to be inscrutable without transparent inner mechanism. To answer this basic question, many studies have been proposed from different aspects, such as features visualization~\cite{feature_visual1}, complex representations disentanglement~\cite{disentangle}, and explainable models~\cite{explainmodel}.

In this work, we focus on the image classification task. Rather than boosting the validation accuracy, we're more interested in the corresponding CNN itself. Some typical examples of visualization and clustering in this work are displayed in Fig.~\ref{fig:intro}. In the context of CNN used for image classification, the subproblems we're trying to explain can be roughly listed as below:
\begin{figure}[t]
\centering
\includegraphics[width=0.95\linewidth]{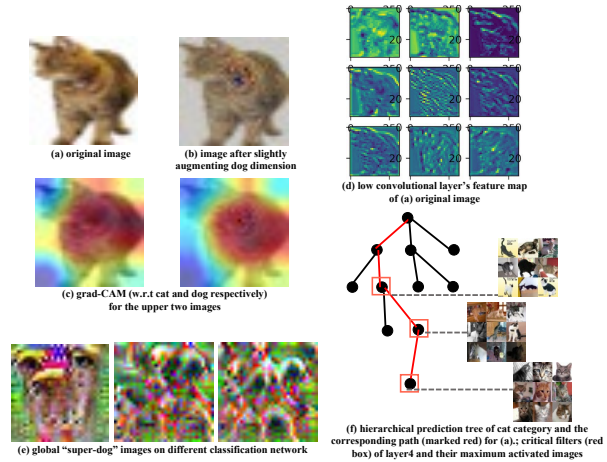}
\caption{Some typical examples of visualization and clustering. (a) is a ground-truth cat image and its feature map of low convolutional layer is (d). By gradient descent on image pixels to augment dog dimension's presoftmax output, (b) is generated and classified as dog. (c) is grad-CAM~\cite{gradcam} showing highly contribution fileds in (a) and (b) for network to predict them as cat and dog, respectively. (e) are the optimized ``dogs'' on different network. (f) is the hierarchical prediction tree for cat category. By querying along the tree and checking filters activation, the prediction path of (a) is marked red, and to explicitly show the semantic features, images max-activating the corresponding critical filters are shown (linked to red box).}
\label{fig:intro}
\end{figure}

\textbf{Neuron knowledge visualization:} With the intuition to open the black box and disentangle the inner knowledge hidden in CNN, a direct and crucial method is to visualize the properties related to a neuron (\emph{a.k.a.} filter) in the CNN~\cite{feature_visual1,feature_visual2,feature_visual3}. The method we adopted is gradient descent \emph{w.r.t.} image pixels and find the input image that maximizes the activation score of some ``units''. More specifically, the term ``units'' can respectively refer to the softmax output (class probability) of a class, the pre-softmax output (class logits), the output value of a middle layer's whole layer, some channel, or a concrete filter. With different optimization goals, the problems tackled are also different:

\begin{itemize}
\item Which features does the network use to recognize and discriminate a class, and how to visualize them? An introducing question would be, how does a little kid, after reading several elephants' pictures and being able to recognize elephants, draw a stick figure of an elephant? The knowledge that a network gains and remembers from training can also be visualized.

\item What features do the low, middle, and high layers in a CNN extract? By sorting the ground-truth images that highly activate a layer and estimate the critical features this layer prefers, we find that convolutional layers in different levels indeed pull out different scope of features.

\item What is exactly activating a specific neuron? Or in a invertible perspective, which regions of the original image is corresponding to which neuron, and how to quantitatively analyze this neurons contribution in the global prediction~\cite{neuroninglobal}? This can be explored by optimizing the single or mutual filters output.
\end{itemize}
\begin{figure}[t]
\centering
\includegraphics[width=0.95\linewidth]{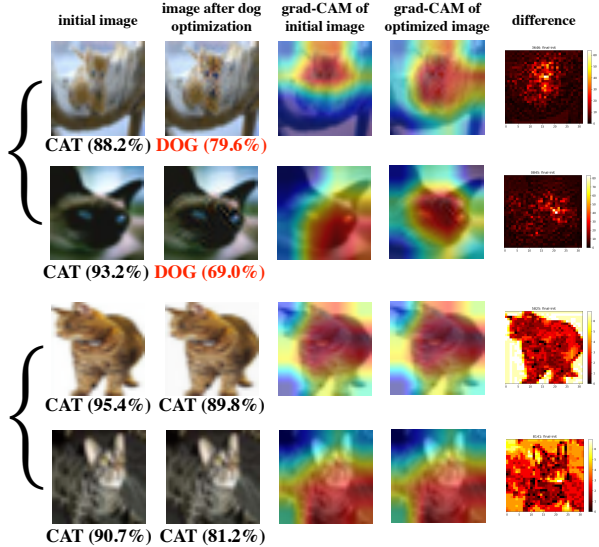}
\caption{Optimize $\Phi(\bx)=$ net.presoftmax$(\bx)_{dog}$, where net is a binary ResNet18~\cite{resnet} trained on dogs and cats of CIFAR-10. Columns are respectively the initial images, optimized images, grad-CAM~\cite{gradcam} of initial images, grad-CAM~\cite{gradcam} of optimized images and difference between optimized and initial images. Rows shows that with different starting images, after a slight gradient descent, the upper two rows are fooled to be dog, while the lower two rows are relatively robust.}
\label{fig:fooling}
\end{figure}

\textbf{Hierarchical clustering:} After we extract and visualize the receptive fields of filters from a local perspective, a natural direction is to dig out in a global view, how does the network combine each filters from discrete features to a final prediction? By clustering the middle layers' feature outputs of images and the corresponding filters, a path can be defined to show how a specific category is predicted along from low-level patterns to high-level compacted feature representations. The clustering algorithm we proposed helps solving the following two problems:

\begin{itemize}
\item What's the hierarchical order among all the categories in images dataset? By the clustering algorithm, the image categories are mapped to a dense graph with subspace distance between them quantitatively defined, and then a hierarchical tree (like taxonomy tree) is generated. This question is challenging and effective in many AI researches: \emph{e.g.} to design a rational and balanced dataset, to classify CNN representations with relationship to biology species.

\item Which filters are responsible for which features, and how much does a filter contribute to the final prediction?  In this research we cluster the filters so that a decision tree is formed, indicating the importance (contribution) of each filters and in which order they're assembled to the final decision. Filters are exactly related to the semantic-level activation over features, and by querying the decision tree of a category from root to leaves, it can be clear how an image is recognized, in an convincing order of semantic fetures from common to specific.
\end{itemize}

\begin{figure*}
\centering
\includegraphics[width=0.95\linewidth]{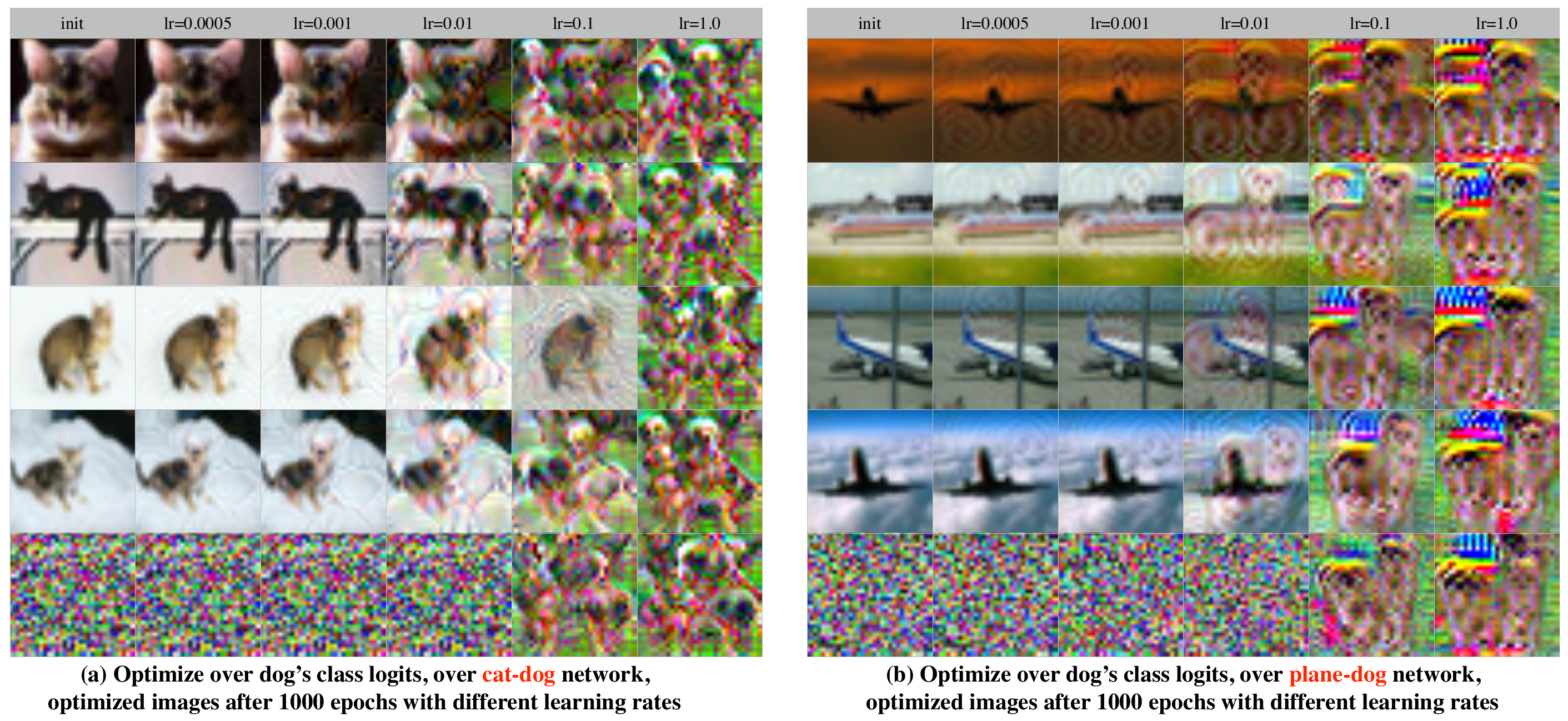}
\caption{Left and right parts show respectively the network trained on cats\&dogs and planes\&dogs, and the optimization objective is the dog dimension of presoftmax output. Initial images are 4 randomly sampled cat (or plane) images from CIFAR-10 and 1 random Gaussian noise with the same channel-wise mean and standard deviation of CIFAR-10 dataset. Columns show respectively the initial images, images after 1000 epochs with different learning rate. The optimized ``dog'' in the left are dog heads and faces, while in the right is a complete dog with body, head, arms and legs.}
\label{fig:catdogplane}
\end{figure*}

\section{Neuron knowledge visualization}

In this section we describe the method of visualising knowledge learned and memorized by a given CNN, some layers or some filters. The correlative experiment results are also attatched in this section. The algorithm is fomulated as reconstructing an optimal image $\bx^*\in\real^{H \times W \times C}$ that maximize some feature output score:
\begin{equation}\label{e:objective}
 \bx^* = \operatornamewithlimits{argmax}_{\bx\in\real^{H \times W \times C}} \Phi(\bx)
\end{equation}
where $\Phi : \real^{H\times W \times C} \rightarrow \real$ indicates the objective to optimize, respectively as a given dimension of the pre-softmax class logits, the average output value of a whole layer or a channel of a layer, the output of a specific filter, or composite sum of the objectives listed above, to examine the collaborative effect of optimizing over multiple goals.

A local optimal $\bx^*$ can be found by back-propagation \emph{w.r.t.} the input image pixels with weights fixed in the pre-trained CNN. As for the starting point $\bx$ of searching, different ground-truth images and the random noise can be chosen. In practice, stochastic gradient descent (SGD) is used to prevent being stuck at a saddle point and reach the global optimal to the full extent~\cite{sgd}.

\subsection{Pixel fooling and robustness of an image}
In the first experiment, a binary classification network is trained using ResNet18 on cat and dog images in CIFAR-10 dataset, achiving validation accuracy of 92.7\%. The interesting intuition is that, how to make a cat look more like a dog? So we optimize the pre-softmax class logits of the dog dimension. Fig.~\ref{fig:fooling} shows the images' change with a low learning rate and training epochs (lr=0.0001, epochs=100).

From the human's visual view the images are modified just slightly and remain the same as cats. However when it comes to network prediction, some of the cat images are misclassified as dog, with a high convincing possibility. By gradient-based localization, the receptive field on original image that highly support the prediction is also marked on grad-CAM~\cite{gradcam} image. The difference $\bdiff$ between final image $\bx_f$ and the starting image $\bx_s$ (gradient accumulated) is also represented in heatmap as $\bdiff_{h,w}=\sqrt{\sum_{c \in C}{|{\bx_f}_{h,w,c}-{\bx_s}_{h,w,c}|^2}}$. 

By observation, images with a messy and complex background are more vulnerable to fooling (if the optimization also considered as fooling). Under the same learning rate and training epochs, these images gain more gradient accumulated and are more likely to be misclassified. The difference image (gradient accumulated) of vulnerable images seems to be randomly sampled, while that of robust images tends to reflect the rough sketch of a cat. Furthermore, the robustness of an image~\cite{foolattack} over a network can be mathematically defined, using terms of difference, learning rate and optimization epochs~\cite{robustness}.

\subsection{Discriminative evidence of a CNN prediction}
Continuing with the cat-dog binary classification network, we still optimize the dog class logits while increasing the learning rate. As shown in Fig.~\ref{fig:catdogplane}, with the increasing learning rate, the visual modification becomes more salient, and when finally the optimized logits converges, many head-shaped patterns appear on the image. One natural guess comes to us that it's because cats and dogs are so similar (both mammal pets, similar body, arms, legs and feather) that the discriminative evidence the network uses to distinguish dogs and cats is their heads. Nowthat we want to reconstruct a ``super-dog'' known by the network, the most discriminative features of dogs (\emph{i.e.} dog head) will appear.

To verify the guess, another binary classification network is trained on plane and dog images in CIFAR-10 dataset, with the same architecture and training hyper-parameters. As shown in Fig.~\ref{fig:catdogplane}, a complete dog (with head, torso, arms and legs) appear from optimization. This maybe because planes and dogs are quite disparate in appearance, thus the network memorize and distinguish a dog by its whole body. The guess we raised above that a CNN's discriminative evidence is related to the training dataset can be roughly verified.

\begin{figure}[t]
\centering
\includegraphics[width=0.95\linewidth]{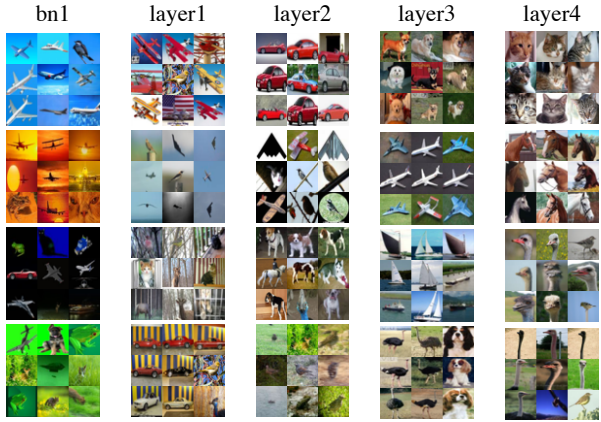}
\caption{Examples of the images that maximize the activation score of some filters in the 1-st batch normalization layer after input transform, 1-st to 4-th convolutional layers. 10-classification network trained whole CIFAR-10. With layer from low to high, filters concern from color, texture to object, part.}
\label{fig:layermaxs}
\end{figure}

\subsection{Optimization in middle layers}
The experiments demonstrated above are optimizing directly the class logits, that is to make something look more like a dog. Then what about just optimizing the middle layers (including channels and neurons)? By doing so the semantic features representation of inner filters will be more clear to us.

First we train a 10-classification network on all the categories of CIFAR-10, using the same architecture as above. Then to evaluate the ground-truth images features on a specific neuron, all the images in dataset are sorted by their respective output of the neuron. Images with the highest activation score of a specific neuron is shown in Fig.~\ref{fig:layermaxs}. In the state-of-art research~\cite{feature_visual1}, semantics for CNN filters can be defined as six types: objects, parts, scenes, textures, materials, and colors. From the sorted images we find that, the high convolutional layers mainly focus on high-level semantics like objects, parts, while the low convolutional layers mainly focus on the low-level features like textures, colors. Here ``low-level'' refers to strong relationship to frequency patterns, as illustrated by some researchers~\cite{fft}, maybe a convolution level indeed acts like a Fourier or some other transform, so that the frequency distribution of the image is concerned.

\begin{figure}[t]
\centering
\includegraphics[width=0.95\linewidth]{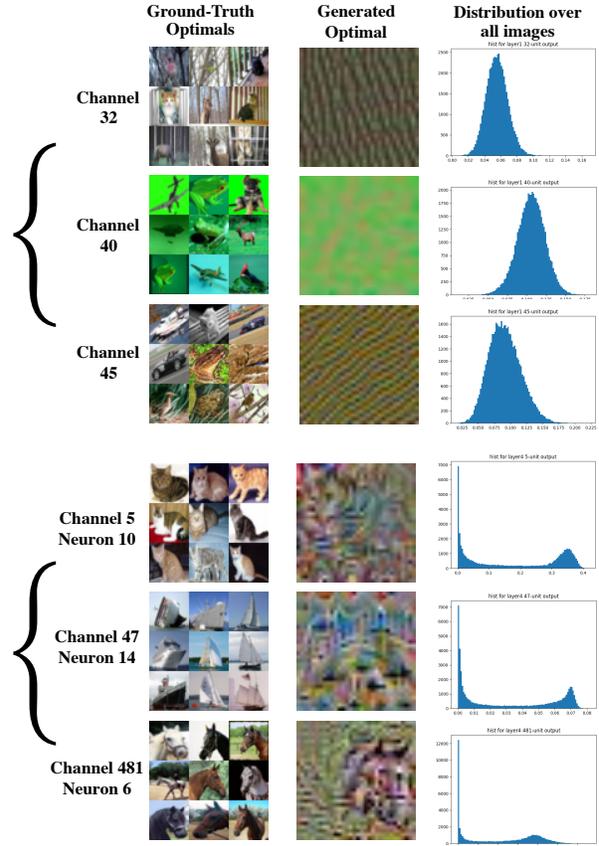}
\caption{Examples of optimizing some specific filters (or channels) in layer1 (upper bracket) and layer4 (lower bracket). Columns show respective the ground-truth optimals, the generated optimals and the activation scores' distribution of all the dataset images.}
\label{fig:layeropti}
\end{figure}

Fig.~\ref{fig:layeropti} shows the result images after the optimization on specific layers or neurons. For the instance of neurons in low convolutional layers, the ground-truth optimals are cage handrails, green background and inclined horizon or branches towards a same direction. The distribution histogram of all the real images' activation score on the filter is bell-shaped, indicating a uniform distribution over the whole dataset. The respective generated optimization are colors and textures. When it comes to the instance of neurons in high convolutional layers, the ground-truth optimals are sailboats, horse heads, and cats sitting in similar posture. The shape of distribution histogram of all the real images' activation score looks much different: only a small number of dataset images can highly activates the filters with high-level semantics of objects, parts, etc., and most of the images are ignored by this filter with low output value.

\begin{figure}[t]
\centering
\includegraphics[width=0.95\linewidth]{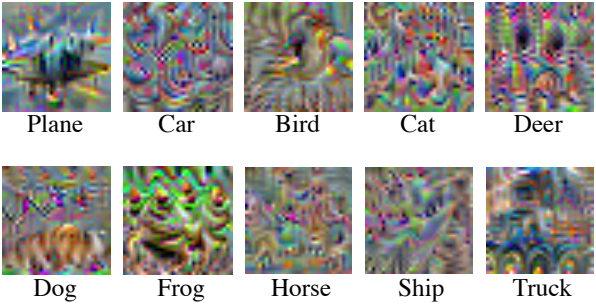}
\caption{``Super'' image examples show how the network memorize and distinguish each category, by optimizing each category's presoftmax output. 10-calssification network trained on whole CIFAR-10 dataset.}
\label{fig:super}
\end{figure}

What's more, we have tried to first generate the ``super'' images Fig.~\ref{fig:super} of categories and then examine whether they bring enough infomation for network classification, by substituting the training set to a small number of ``super'' images. The result is not good. Mathematically it's like back-stepping from a local optimal point to the whole function, and the infomation entropy is increased out of thin air thus it's impossible~\cite{entropy}. Another direction worth trying would be add restrictions to preserve entropy, or fine-tune just the high convolutional layers in a pre-trained network.

\subsection{How to make the generated images look more lifelike?}

In the generated optimal images shown above, they're filled with high frequency noise and nail-shaped patterns. These nonsensical patterns are partly because of the pooling operation in forward process, and during back-propagation, a block of pixels are motivated simultaneously to respond the network.

\begin{figure}[t]
\centering
\includegraphics[width=0.85\linewidth]{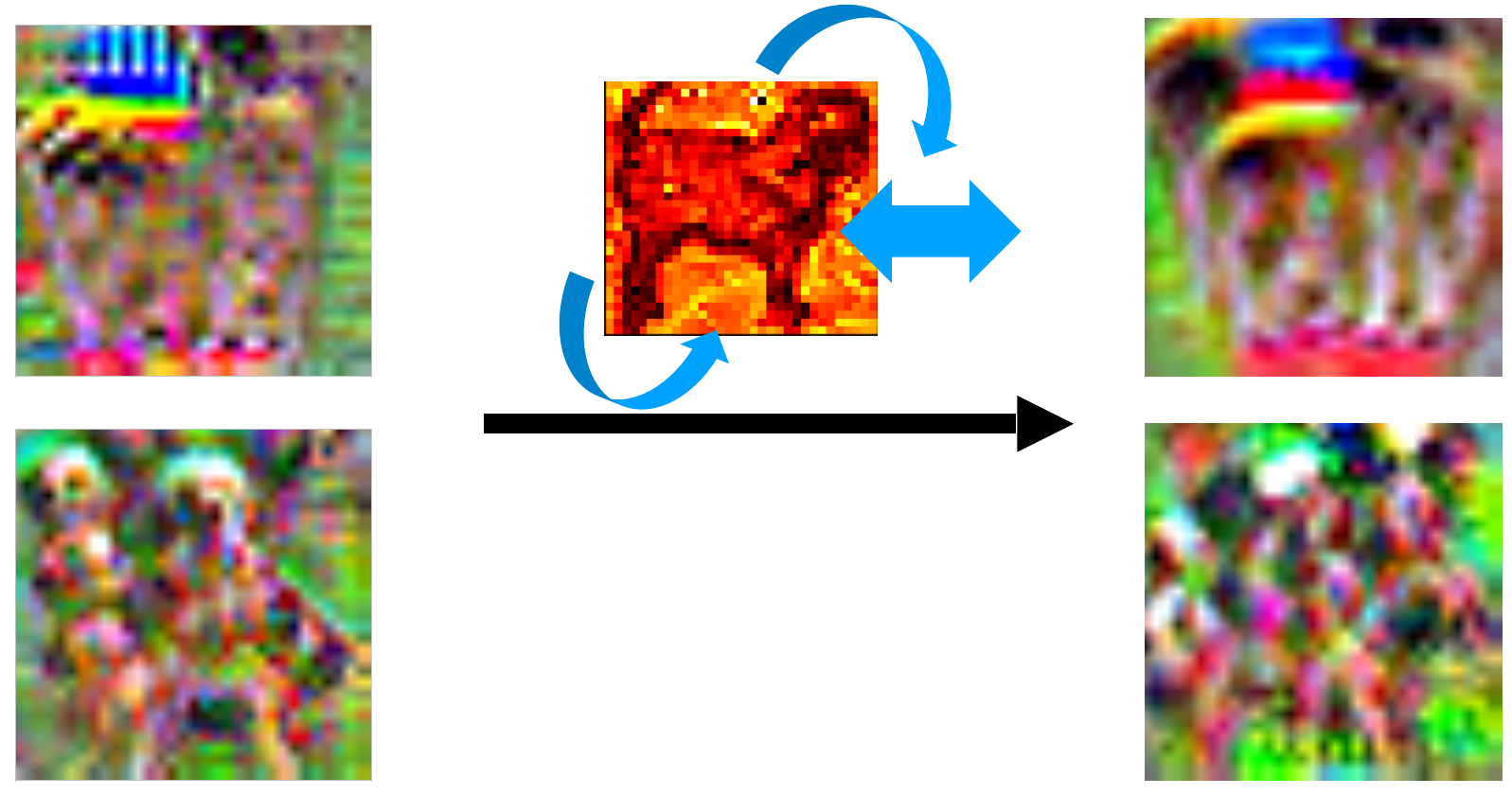}
\caption{Optimization examples before and after adopting the trick that randomly blur, translate and rotate the gradient.}
\label{fig:lifelike}
\end{figure}

To eliminate these patterns and make the generated images look more lifelike, here I use the trick that, every time the gradient is back propagated, before subtracted by the original image, it's randomly blurred, translated and rotated in a slight extent~\cite{addnoise}. The result is shown in Fig.~\ref{fig:lifelike}.

Another way is inspired by ~\cite{regulate}: to add a regulariser term to the optimization objective in \eqref{e:objective}, \emph{s.t.}
\begin{equation}\label{e:objective2}
 \bx^* = \operatornamewithlimits{argmax}_{\bx\in\real^{H \times W \times C}} {(\Phi(\bx)- \lambda \mathcal{R}(\bx))}
\end{equation}
where $\lambda >0$ is the regularizer coefficient and $\mathcal{R} : \real^{H \times W \times C} \rightarrow \real$ is a regulariser to measure how lifelike the image $\bx$ is. The lower value of $\mathcal{R}(\bx)$, the more likely $\bx$ looks to be natural and lifelike.

The design of regulariser exactly depends on how to define ``natural'': The simple version would be the $\alpha$-norm  $\mathcal{R}_\alpha(\bx) = \|\bx\|_\alpha^\alpha$, where $\bx$ is the vectorised and mean-subtracted image. With a relatively large $\alpha$ (\emph{e.g.} 6) and subtracting-mean operation, this regulariser encourage the images to be constrained within an interval rather than diverging.

Another useful regulariser is proposed as total variation to measure the variation extent in neighbor pixels. For each discrete channel ($\bx \in \real^{H \times W}$) of the original image, total variation can be approximated by the finite-difference:
$$
 \mathcal{R}_\TV(\bx)
 =
 \sum_{i,j}
 \left(
 \left(x_{i,j+1} - x_{ij}\right)^2 +
 \left(x_{i+1,j} - x_{ij}\right)^2
 \right)^\frac{\beta}{2}.
$$
When $\beta=1$ it's exactly the presence times of subsampling, which also caused by max pooling in CNN, while max pooling indeed leads to the high-frequency patterns.

\subsection{Out-of-sample prediction}
Which category would an image not belonging to the trained categories be predicted as? Initially we expect that untrained images would be randomly sampled into each category, however it turns out not exactly: as shown in Table~\ref{tab:outofsample}, in a binary classification network distinguishing cats and dogs, except for horse (72.5\% of the horse images are classified as dog), all the other out-of-sample images are mostly classified as cat, and for the random noise images, even all of them are recognized as cat. The similar phenomenon also happens to plane-dog network and the 10-classification network: nearly all the random noise images are recognized as bird.

\begin{table}[htbp]
\centering
\resizebox{0.9\linewidth}{!}{\begin{tabular}{c|cccccccc|c}
&\!\! Plane \!\!&\!\! Car \!\!&\!\! Bird \!\!&\!\! Deer \!\!&\!\! Frog \!\!&\!\! Horse \!\!&\!\! Ship \!\!&\!\! Truck \!\!&\!\! Random Noise\\
\hline
Cat(\%) & 91.9 & 81.4 & 70.1 & 59.1 & 85.7 & & 93.1 & 73.8 & 100.0\\
Dog(\%) &  &  &  &  &  & 72.5 &  &  & \\
\end{tabular}}
\vspace{2pt}

\centering
\resizebox{0.9\linewidth}{!}{\begin{tabular}{c|cccccccc|c}
&\!\!  Car \!\!&\!\! Bird \!\!&\!\! Cat \!\!&\!\! Deer \!\!&\!\! Frog \!\!&\!\! Horse \!\!&\!\! Ship \!\!&\!\! Truck \!\!&\!\! Random Noise\\
\hline
Plane(\%) & 88.8 &  &  &  &  & & 95.1 & 86.0 & 99.7\\
Dog(\%) & & 63.3 & 90.9 & 81.1 & 74.8 & 85.6 &  &  &  \\
\end{tabular}}
\vspace{2pt}
\caption{Percentage of out-of-sample images' prediction to some category, network trained respectively on cat-dog and plane-dog in CIFAR-10, and test images also from CIFAR-10}
\label{tab:outofsample}
\end{table}

What we're concerned about is not to alleviate the problem, where some effective solutions already exist~\cite{oosample}, but to make clear the mathematical reason why, say, random images are all recognized as bird.

The intuitive guess is about bias after fully connected layer: CNN filters identify ``features'' and issues the presence or lack of them to determine which class the image belongs to. Due to the lack of salient physical features, random images cannot activate any of the filters, thus $\by=\bw^T\bx+b \approx b$, and they'll mostly come down to the highest bias term. However, the bird dimension of bias isn't the highest, and what's more, the $\bw^T\bx$ term for random noise $\bx$ isn't small.

Then a question comes to us: whether the property carry over to different architectures? \emph{I.e.} is it an incidental property of the network or is it a property of the dataset. We re-train the network with different random initial weight, and train some other architectures like VGG-16, and still, all the random images are classified as bird. Thus the phenomenon tends to be dataset-specific~\cite{cifar10}.

So what property of bird images in CIFAR-10 dataset exactly have? Mathematically the discrimination range (\emph{i.e.} a category's images being ``clean/dirty''~\cite{cleandirty}) can be measured by the spanning space: the bird images' subspace spans with too large a radius that covers the random image space. The K-Means algorithm can be used to measure the cluster range, and the metric to name the clusters as categories for measuring distance can be exactly fomulated as a stable marriage problem, where Gale-Shapley algorithm works. However in original images clusters, bird images don't show anything special.

Another assumption is the dataset bias~\cite{datasetbias}: for an extreme example, in all the bird images the bird appears with a random noise background. Nevertheless, this assumption still needs to be verified, and we would more like to firstly dig out and cluster the distribution of images space - this directly leads to the next section of our research: clustering.

\section{Hierarchical clustering}
In this section, we design two clustering algorithm to cluster respectively over image categories and CNN filters. The first algorithm aims to define and calculate the distance among categories in images space, and furthermore generate a hierarchical classification tree. The second algorithm basically focuses on how the filters are associated with different hierarchical priority to produce the final category prediction, thus a filter-wise decision tree is generated.

\subsection{Unsupervised category clustering}
When decoding an image category into distribution in images space, one natural observation is that hierarchy exists in categories: for example in CIFAR-10 dataset, ten categories can be divided into two super-categories: animals and vehicles:
\includegraphics[width=0.9*\linewidth]{tree.tikz}
Then with this super-category clustering, some further work like hierarchical learning, new category network refinement can be achieved. However, this classify approach is with prior knowledge and not precise enough: suppose there're distance $d$ in space among categories, for super-categories $S,\ T$ and for basic categories $\forall A,B,C \in S, \ D\in T,$ the method is based on the assumption that $d(A,B)\approx d(A,C)<d(A,D)$. Nevertheless, one may say that $d($dog,cat$)<d($dog,horse$)$, or even $d($bird,plane$)<d($bird,frog$)$.

Taking this circumstance into consideration, we propose the unsupervised category clustering: without human's prior knowledge, the categories are unsupervised clustered using the knowledge learned in the network.

\begin{algorithm} [t] 
\caption{unsupervised learning for the hierarchical structure for categories} 
\label{algA}
    \LinesNotNumbered
    \SetKwFor{For}{for}{do}{}
    \SetKwFor{If}{if}{then}{}
    \SetKwInOut{Input}{input} \SetKwInOut{Output}{output}
    % \SetKwInput{kwInit}{Initialization}  \kwInit{}
    \Input{ a pre-trained CNN,\\image set ${\Omega}=\{~\Omega_c~|$for $c$ in categories set $C\}$}
    \Output{ The hierarchical structure of categories}
    $G$=Graph()\;
    $\mu_c$=representitive vector for a category $c$ \;
    \For{$\Omega_c~\in~\Omega$}{
    	$\mu_c\leftarrow\overline{\mu_{c_i}}$, where $\mu_{c_i}$=feature map of a given layer of $I_{c_i}$, \textbf{for} each image $I_{c_i} \in \Omega_c$}

\For{$\forall c_1,c_2 \in C$}{$G$.addEdge($c_1,c_2$,weight=$\frac{1}{2}$(1-$cos(\mu_{c_1},\mu_{c_2})))$}

\While{\text{\upshape in $G$ more than one nodes don't have a parent node}}{
$c_u,c_v$ $\leftarrow$  $\mathop{\arg\min}\limits_{c_u,c_v \in G} \{~|(c_u,c_v)|~$\big|$~G$.existEdge$(c_u,c_v)$\\$\>\>\>\>\>\>\>\>\>\>\>\>\>\>\>\>\>\>\>\>\>\>\>\>\>\>\>\>\>\>\>\>\>\>\>$and~$G$.notParentChild$(c_u,c_v)\}$ \;
add a supernode $s$ for $c_u$ and $c_v$ \;
$G$.addParentChildEdge($s,c_u$) \;
$G$.addParentChildEdge($s,c_v$) \;
$G$.removeEdge($c_u,c_v$)\;
\For{ $c_w \in$ G.\text{\upshape nodes} ${\backslash\{c_u,c_v,s,c_u.\text{\upshape children},c_v.\text{\upshape children}\} }$ }{
    \If{\text{\upshape $G$.existEdge($c_w,c_u$) or $G$.existEdge($c_w,c_v$)}}{
        $G$.addEdge($c_w,s$,weight=$\overline{|(c_w,c_m)|}$ \textbf{for} $c_m$ in \{$c_u,c_v$\}) \;
        $G$.removeEdge($c_w,c_u$)\;
        $G$.removeEdge($c_w,c_v$)\;
    }
}
}
\end{algorithm}

The pseudocodes are illustrated in Algorithm~\ref{algA}. The first issue is how to represent and calculate the distance among categories. Here the vectorised output of the 4-th layer (last convolutional layer before fully connected layer) in the ResNet18 network is adopted to represent a image's feature. Then for each category, its representative vector is the average vector of all the corresponding vectors of images under this category. To measure the distance between each two categories, we simple calculate the cosine value between the two corresponding representative vectors~\cite{cosine}.

With definition of representative vectors and category distance, a dense graph can be constructed, with each node representing a category, and the edge linking some two nodes is assigned with the weight of distance between the two categories.

To generate the hierarchical structure of categories, the algorithm uses the greedy method like Boruvka algorithm~\cite{boruvka}: every time the shortest non-parent-child edge is picked up and cut off, and a parent super node is created and linked to the edge's two endnodes. For any other nodes linked to some of the two endnodes, the edges are cut off and re-linked to the parent super node, with edge weight assigned to be average distance to the two endnodes. Iterate until the hierarchical structure is completely generated. With this algorithm the Minimum Spanning Tree (MST) of the original dense graph can also be generated.

\begin{figure}[t]
\centering
\includegraphics[width=0.85\linewidth]{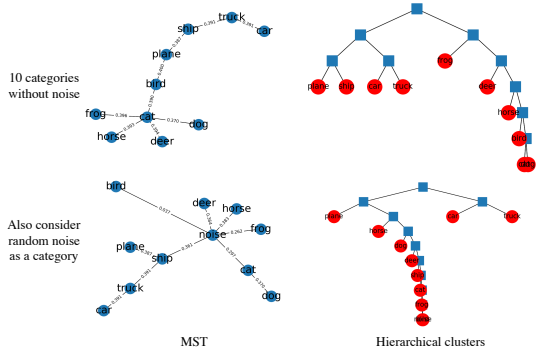}
\caption{Examples of the generated MST and hierarchical clusters. Netowrk trained on whole CIFAR-10 dataset and the last convolutional layer's output is used as representative vector for each category. The upper and lower case show whether or not to include random noise as an independent category.}
\label{fig:hier}
\end{figure}
\begin{figure}[t]
\centering
\includegraphics[width=0.85\linewidth]{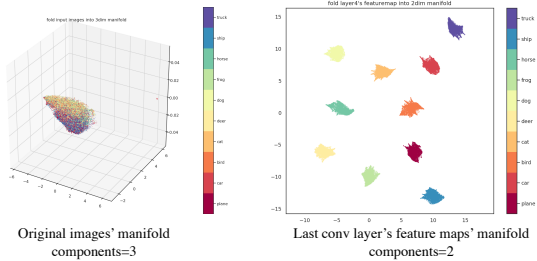}
\caption{Using UMAP algorithm~\cite{umap} to reduce dimension of respectively the original images and the representative vectors. The original images' manifold in 3-dimension is still overlapped messily, while representative vectors' manifold in 2-dimension is divided and clustered.}
\label{fig:manifold}
\end{figure}

From Fig.~\ref{fig:hier} see the MST and hierarchical structure of CIFAR-10 categories for example: in the instance of only 10 categories, cat is a center in MST, which can partly explain the dataset biases mentioned above: many out-of-sample images are predicted as cat. The distance between bird and plane is exactly rather small, indicating their similarity in a way. As for the hierarchical clusters, the cat and dog are in the lowest clustering level, and while stepping upward, categories are classified to two big categories, while different real categories are in different level (from root to leaves, \emph{i.e.} from rough to fine-grained.) Then we come to another case if random noise is also considered as an independent category: noise tends to be a centroid in MST, and the distance between bird and noise is the smallest.

The category distance defined here can be multiplied to the original loss function as a cross-category penalty in online training. Also we would like to expand the hierarchical distance algorithm to multi object classification \& detection task (one filter not restricted in only one category).

What's more, to make sure the representative vectors used here bring enough infomation for discrimination, all the images' original self-flatten vectors and the corresponding representative vectors are folded from subspace into a low dimension sphere using the Uniform Manifold Approximation and Projection algorithm~\cite{umap}. Fig.~\ref{fig:manifold} shows that the last convolutional layer's output is indeed highly representative for image classification.

\begin{figure*}
\centering
\includegraphics[width=0.95\linewidth]{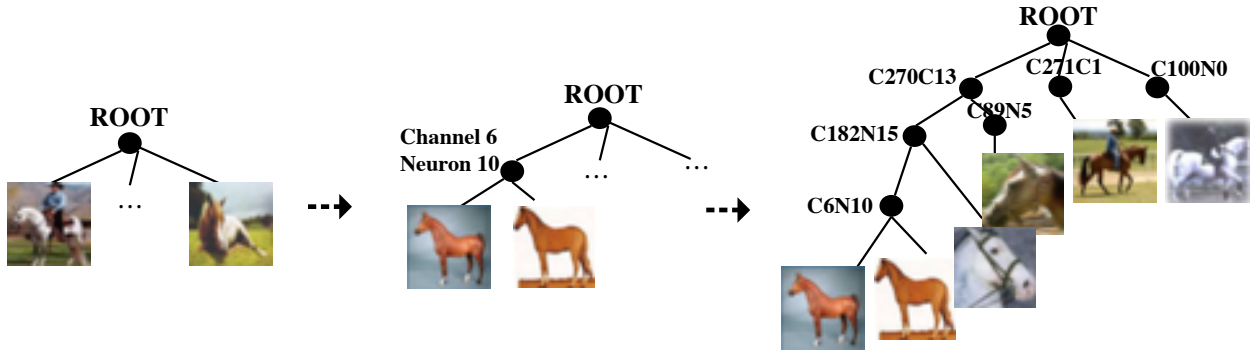}
\caption{Process of using hierarchical cluster algorithm to generate a prediction tree. Use horse category as an example: each leaf node represents a horse image that drops down to the final prediction slot. By stepping and merging upwards to root, different nodes implicitly represent different semantic features from fine-grained to rough (\emph{e.g.} standing posture, mane, neck, head, etc.), and each node has its critical filter, which contributes most in the node's semantic representation.}
\label{fig:horsetree}
\end{figure*}

\subsection{Filter clustering: prediction tree}
\begin{algorithm} [t]
\caption{pull out a category's filter-wise hierarchical tree of a specific layer}
\label{algB}
    \LinesNotNumbered
    \SetKwFor{For}{for}{do}{}
    \SetKwFor{If}{if}{then}{}
    \SetKwInOut{Input}{input} \SetKwInOut{Output}{output}
    % \SetKwInput{kwInit}{Initialization}  \kwInit{}
    \Input{ A given layer $L$ with $d$ filters in a pre-traind CNN, $L$'s feature maps $\Omega\in \mathbb{R}^{c\times d\times l \times l} $ of $c$ images in the category}
    \Output{ the layer $L$'s filter-wise hierarchical tree over the category}
    create a tree $T$, with root $R$, edegs ($R,\Omega_i$) for all $\Omega_i \in \mathbb{R}^{d\times l \times l}$ in $\Omega$;\\ 

    $R$.filters = $\Omega_i$.filters = $[1,\ldots,d]$;

    \While{$\exists$ \text{\upshape a child} $p$ \text{\upshape of} $R$ s.t. $|p.\text{\upshape filters}|>1$}{
        $u,v$ $\leftarrow$ $\mathop{\arg\min}\limits_{u,v \in R.\text{\upshape childern}} \{ \frac{1}{2}(1-cos(u,v)) \}$ \;

        add a supernode $s=\overline{u,v}$ linked with $u,v$, disconnect $u,v$ from $R$, connect $s$ to $R$\;
        
        critical filter $F$ $\leftarrow$  $\mathop{\arg\min}\limits_{ F \in (u.\text{\upshape filters} \cap  v.\text{\upshape filters})}\frac{cos(u \odot \mathbbm{1}_{d}^{u.\text{\upshape filters}\backslash \{F\}},v \odot \mathbbm{1}_{d}^{v.\text{\upshape filters}\backslash\{F\}})}{cos(u \odot \mathbbm{1}_{d}^{u.\text{\upshape filters}},v \odot \mathbbm{1}_{d}^{v.\text{\upshape filters}})}$  \;
        //$\mathbbm{1}_{d}^{M}=[1 \ \text{\upshape if} \ i \in M \ \text{\upshape else}\ 0 \ \textbf{for}\ i \in 1,\ldots,d]^{\textbf{T}}$, $\odot$ is point-wise multiplication;\\
        s.filters=\{$u.\text{\upshape filters} \cap v.\text{\upshape filters}$\}$\backslash$ $\{F\}$ \;
    }
     
\end{algorithm}

This algorithm can be viewed as an extension to the Algorithm~\ref{algA} to hierarchically cluster categories: we go deeper into only one specific category, and cluster the filters of a given layer and then finally generate a filter-wise hierarchical tree that defines along which path an image is recognized into this category.

The tree is constructed where each node represents a semantic feature (unsupervised clustered, so without a specific name assigned, need afterwards annotation by induction on images features). Foe each node in the tree, it's associated with some critical filters, by storing the attribute of the filters that ``reachable'' from root to it.

See pseudocodes are illustrated in Algorithm~\ref{algB} for detailed implementation: initially the representative vectors of all the images under this category are linked as child node to the root, with reachable filters all set to whole filters. Then from the root's children pick up two nodes that are closest among all the children nodes, merge the two nodes into one super node, which means a more common level of semantic feature. From the intersection of the two nodes' reachable filters, a critical filter is defined as the filter without which the cosine distance between the two nodes would be changed mostly, then the reachable filters of the super node is assigned as two nodes' reachable filters' intersection with removal of the critical filter. Repeat the iteration until all the root's children only have one reachable filter.

This algorithm cluster the filters and generate the filter-wise hierarchical tree. The tree represents the decision path of the given category: with each node representing one semantic feature, it's critical filter is the difference of it's reachable filters set to the intersection of its two children nodes' reachable filters set.

Taking the horse category for example (see Fig.~\ref{fig:horsetree}), from the root to leaves the features are clustered in the order from rough to fine-grained: some filters first judge the presence of the horse head or the horse torso, then along the way to the presence of horse mane, neck, etc., examine the direction of horse standing posture, and finally leads to a leaf, representing one final prediction mode.

The application of filter-wise hierarchical cluster is potential: for example, next time a new image is classified as a horse, we can search along the tree to examine which filters are activated and see clearly which patterns of the image highly contribute to the final prediction. What's more, consider an application scenario with onerous dataset and complex network where it consumes a lot to train from the sketch. If some new images are added to the dataset and the network need to be updated online~\cite{online} (\emph{i.e.} the mass face recognition system), it's not necessary to re-trained the network, but just figure out the prediction paths of the new images, and then only fine-tune the weights related to the critical filters in the paths.

For further work, we would like to expand the algorithm so that it's not restricted in one category, but to the whole image space - the inner filters don't know about final categories, but just pull out the specific features. With expansion to the whole images, the feature representation of filters can be more comprehensive.
\section{Conclusion and discussions}

In this study to explore what neurons in CNN learn, we use methods of visualize and cluster.

By visualizing the knowledge hidden in middle layers, we find that different images are robust in different extent over a specific network, which gives inspiration to adversarial attacking design. We also find that the discrimination evidence the network memorized is related to training set distribution.

In the section of cluster, the algorithms respectively work for hierarchically classifing the image categories and generating the prediction tree over filters. The representative vectors used in the first aspect prove to be effective and the level of categories coincides to our intuition. For the second aspect, we find that different filters extract different features, with low layers concerning like textures and high layers concerning like parts. Even for filters in a same layer, they are exactly in different hierarchy in attributing to the final prediction.

However we also need to note that a basic question that ``why random images are all predicted as bird'' hasn't been answered reasonably, thus deeper research need to be conducted \emph{w.r.t.} category's space. Another issue is that what if the filters don't learn what human call ``features'' but learn over not-visual-salient features like frequency distribution of images, and we need to visualize features of manifolds of images, rather than features of images. What's more, the clustering algorithm over filters can only produce the approximate order of filters attribution. For further work we would like to quantitatively define how much a filter's activation attributes to the final prediction.
{\small
\bibliographystyle{ieee}
\bibliography{TheBib}

\begin{thebibliography}{10}\itemsep=-1pt

\bibitem{learnquestion}
M.~T. Ribeiro, S.~Singh, and C.~Guestrin.
\newblock ``why should i trust you?'' explaining the predictions of any
  classifier.
\newblock {\em In KDD}, 2016.

\bibitem{feature_visual1}
D.~Bau, B.~Zhou, A.~Khosla, A.~Oliva, and A.~Torralba.
\newblock Network dissection: Quantifying interpretability of deep visual
  representations.
\newblock {\em In CVPR}, 2017.

\bibitem{feature_visual2}
C.~Szegedy, W.~Zaremba, I.~Sutskever, J.~Bruna, D.~Erhan, I.~Goodfellow, and R.~Fergus.
\newblock Intriguing properties of neural networks.
\newblock {\em In {arXiv:1312.6199v4}}, 2014.

\bibitem{feature_visual3}
M.~D. Zeiler and R.~Fergus.
\newblock Visualizing and understanding convolutional networks.
\newblock {\em In {ECCV}}, 2014.

\bibitem{disentangle}
Q.~Zhang, R.~Cao, Y.~N. Wu, and S.-C. Zhu.
\newblock Growing interpretable part graphs on convnets via multi-shot learning.
\newblock {\em In {AAAI}}, 2016.

\bibitem{explainmodel}
Q.~Zhang, R.~Cao, F.~Shi, Y.~Wu, and S.-C. Zhu.
\newblock Interpreting cnn knowledge via an explanatory graph.
\newblock {\em In AAAI}, 2018.

\bibitem{gradcam}
R.~R. Selvaraju, M.~Cogswell, A.~Das, R.~Vedantam, D.~Parikh, and D.~Batra.
\newblock Grad-cam: Visual explanations from deep networks via gradient-based
  localization.
\newblock {\em In arXiv:1610.02391v3}, 2017.

\bibitem{neuroninglobal}
S.~M. Lundberg and S.-I. Lee.
\newblock A unified approach to interpreting model predictions.
\newblock {\em In NIPS}, 2017.

\bibitem{sgd}
A.~Brutzkus and A.~Globerson.
\newblock Globally optimal gradient descent for a ConvNet with Gaussian inputs
\newblock {\em In arXiv:1702.07966}, 2017.

\bibitem{robustness}
A.~Fawzi, O.~Fawzi, and P.~Frossard.
\newblock Analysis of classifiers' robustness to adversarial perturbations.
\newblock {\em In Mach Learn 107, 481–508, doi:10.1007/s10994-017-5663-3}, 2018.

\bibitem{resnet}
K.~He, X.~Zhang, S.~Ren, and J.~Sun.
\newblock Deep residual learning for image recognition.
\newblock {\em In {CVPR}}, 2016.

\bibitem{cifar10}
B.~Recht, R.~Roelofs, L.~Schmidt, and V.~Shankar.
\newblock Do CIFAR-10 classifiers generalize to CIFAR-10?
\newblock {\em In arXiv:1806.00451}, 2018.

\bibitem{foolattack}
J.~Su, D.~V. Vargas, and S.~Kouichi.
\newblock One pixel attack for fooling deep neural networks.
\newblock {\em In arXiv:1710.08864}, 2017.

\bibitem{fft}
Y.~Wang, C.~Xu, S.~You, D.~Tao, and C.~Xu.
\newblock CNNpack: Packing convolutional neural networks in the frequency domain.
\newblock {\em In {NIPS}}, 2016.

\bibitem{entropy}
M.~Gabrié, A.~Manoel, C.~Luneau, J.~Barbier, N.~Macris, F.~Krzakala, and L.~Zdeborová.
\newblock Entropy and mutual information in models of deep neural networks.
\newblock {\em In {NIPS}}, 2018.

\bibitem{addnoise}
D.~Smilkov, N.~Thorat, B.~Kim, F.~Viégas, and M.~Wattenberg.
\newblock SmoothGrad: removing noise by adding noise.
\newblock {\em In arXiv:1706.03825}, 2017.

\bibitem{regulate}
A.~Mahendran and A.~Vedaldi.
\newblock Understanding deep image representations by inverting them.
\newblock {\em In {CVPR}}, 2015.

\bibitem{oosample}
Kimin.~Lee, Kibok.~Lee, H.~Lee, and J.~Shin.
\newblock A simple unified framework for detecting out-of-distribution samples and adversarial attacks.
\newblock {\em In {NIPS}}, 2018.

\bibitem{cleandirty}
H.~Wickham.
\newblock "Tidy Data".
\newblock {\em in Journal of Statistical Software, Foundation for Open Access Statistics}, vol. 59(i10), 2014.

\bibitem{datasetbias}
Q.~Zhang, W.~Wang, and S.-C. Zhu.
\newblock Examining cnn representations with respect to dataset bias.
\newblock {\em In {AAAI}}, 2018.

\bibitem{cosine}
C.~Luo, J.~Zhan, L.~Wang, Q.~Yang.
\newblock Cosine normalization: Using cosine similarity instead of dot product in neural networks.
\newblock {\em In arXiv:1702.05870v5}, 2017.

\bibitem{boruvka}
Borůvka and Otakar.
\newblock "O jistém problému minimálním" [About a certain minimal problem].
\newblock {\em In Práce Mor. Přírodověd. Spol. V Brně III (in Czech and German)}, 1926.

\bibitem{umap}
L.~McInnes, J.~Healy, and J.~Melville.
\newblock UMAP: Uniform manifold approximation and projection for dimension reduction.
\newblock {\em In arXiv:1802.03426v2}, 2018.

\bibitem{online}
D.~Sahoo, Q.~Pham, J.~Lu, and S.~C.H. Hoi
\newblock Online deep learning: Learning deep neural networks on the Fly.
\newblock {\em In arXiv:1711.03705}, 2017.

\end{thebibliography}
}

\end{document}